\pdfoutput=1

\documentclass[11pt]{article}

\usepackage{acl}

\usepackage{times}
\usepackage{latexsym}

\usepackage[T1]{fontenc}

\usepackage[utf8]{inputenc}

\usepackage{microtype}

\usepackage{amsfonts}
\usepackage{amsmath}
\usepackage{graphicx}
\usepackage{multirow}
\usepackage{caption}
\usepackage{subfigure}
\usepackage{enumitem}
\usepackage{float}
\usepackage{makecell}
\usepackage{booktabs}
\usepackage{tabu}

\newcommand{\ours}{HVPNeT}

\usepackage{booktabs}
\usepackage{amssymb}
\usepackage{pifont}
\newcommand{\xmark}{\ding{55}}%

\newtheorem{remark}{\noindent \textbf{Remark}}

\title{

\emph{Good Visual Guidance Makes A Better Extractor:} \\Hierarchical Visual Prefix for Multimodal Entity and Relation Extraction}

\author{
	Xiang Chen$^{1,2}$, Ningyu Zhang$^{1,2}$\footnotemark[1] , Lei Li$^{1,2}$,
	Yunzhi Yao$^{1,2}$, Shumin Deng$^{1,2}$, \\
	\textbf{
	Chuanqi Tan$^{3}$, 
	Fei Huang$^{3}$, Luo Si$^{3}$,
	Huajun Chen$^{1,2,}$\thanks{$\quad$ Corresponding Author.} 
	} \vspace{1.0mm} \\
	$^1$Zhejiang University \& AZFT Joint Lab for Knowledge Engine, China \\
	$^2$Hangzhou Innovation Center, Zhejiang University, China \\
	$^3$Alibaba Group, China \\
	\fontsize{11}{10}\selectfont 
	\{xiang\_chen, zhangningyu, leili21, yyztodd, 231sm, huajunsir\}@zju.edu.cn,  \\
	\fontsize{11}{10}\selectfont \{chuanqi.tcq, f.huang, luo.si\}@alibaba-inc.com \\
}

\begin{document}
\maketitle
\begin{abstract}
Multimodal named entity recognition and relation extraction (MNER and MRE) is a fundamental and crucial branch in information extraction. However, existing approaches for MNER and MRE usually suffer from error sensitivity when irrelevant object images incorporated in texts. To deal with these issues, we propose a novel \textbf{H}ierarchical \textbf{V}isual \textbf{P}refix fusion \textbf{NeT}work (\textbf{HVPNeT}) for visual-enhanced entity and relation extraction, aiming to achieve more effective and robust performance. Specifically, we regard visual representation as pluggable visual prefix to guide  the textual representation for error insensitive forecasting decision. We further propose a dynamic gated aggregation strategy to achieve hierarchical multi-scaled visual features as visual prefix for fusion. Extensive  experiments on three benchmark datasets demonstrate the effectiveness of our method, and achieve  state-of-the-art performance\footnote{Code is available in \url{https://github.com/zjunlp/HVPNeT}.}.

\end{abstract}

\section{Introduction}

Named entity recognition (NER) and relation extraction (RE) are important tasks in information extraction and knowledge base population, due to its research significance in natural language processing (NLP) and wide applications \cite{hosseini2019,DBLP:conf/emnlp/ZhangDBYYCHZC20,erica,DBLP:journals/corr/abs-2106-01686}. 
Currently, with the rapid development of multimodal learning, multimodal NER (MNER) and Multimodal RE (MRE) methods~\cite{moon2018multimodal,multimodal-re} have been proposed to enhance linguistic representations with the aid of visual clues from images.
It significantly extends the text-based models by taking images as additional inputs, since the visual contexts help to resolve ambiguous multi-sense words.

\begin{figure}[!htbp] 
\centering 
\includegraphics[width=0.45\textwidth]{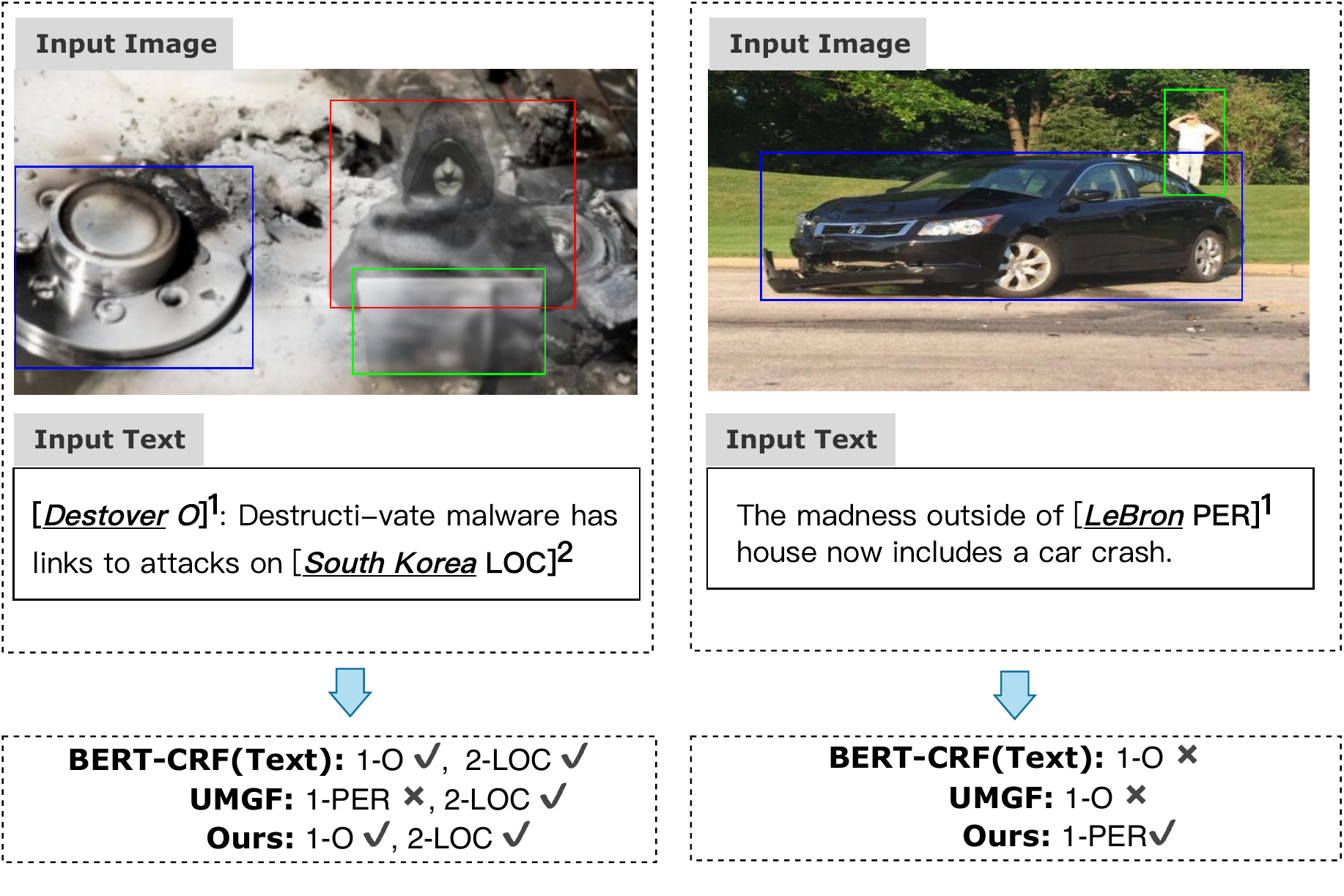} 
\caption{Motivation for robust and effective hierarchical modality fusion.} 
\label{fig:motivation}
\end{figure}

The essence of MNER and MRE tasks is how to learn great visual features and how to incorporate it into textual representation for enhancing NER and RE.
Early methods \cite{zhang2018adaptive,moon2018multimodal} study how to incorporate the feature of whole image into the textual representation. 
\citet{yu-etal-2020-improving,zhang-UMGF,multimodal-re} further validate that  object-level visual fusion is more specific and important for MNER and MRE.
Recently,  RpBERT~\cite{sun2021rpbert} propose to train a classifier of whether the “Image adds to the tweet meaning” before MNER tasks.
However, they heavily rely on pre-training on large extra annotated corpus of image-text relevance and 
only focus on the  whole image with ignoring the bias of relevant object-level visual fusion. 
In practice, irrelevant objects may directly exert negative effects on the text inference. 
Meanwhile, it is not trivial to acquire absolutely relevant object-level visual information to enhance the text.
Thus, an effective method should be derived to learn better visual representation and alleviate error sensitivity of irrelevant object images for social media NER and RE tasks.

Considering images often appear before the text in a web document, we argue that images can be regarded as the prefix for their textual descriptions, which is inspired by prompt learning~\cite{lmbff,prefix-tuning,DBLP:journals/corr/abs-2204-04392,DBLP:journals/corr/abs-2108-13161} in the language model.
Specifically,  given a image-text pair, we prepend object-level image feature sequence of length $V_i$ (visual prefix) to the text sequence at each self-attention layer of BERT~\cite{devlin2018bert}. Note that the visual prefix is a pluggable operation and don't require any annotation on relevance.
Therefore, visual prefix can not only introduce object-level visual signals, but also further reduce the impact on the architecture representing text. Intuitively, visual prefix regarded as a prompt for text may helps alleviate the error sensitivity of irrelevant object images.

While Convolution Neural Networks (CNNs) contain the multi-scale information with pyramidal feature hierarchy~\cite{ren2015faster} from low to high levels. And BERT  encodes a rich hierarchy of linguistic information~\cite{anna2019} from the bottom to the top.
Inspired by \citet{lin2017feature,panet}  that objects of different sizes can have appropriate feature representations at the corresponding scales, we propose to make each layer of BERT aware of hierarchical multi-scale visual features to make a more enlightened and comprehensive forecasting decision.

To this end,
we propose a novel \textbf{H}ierarchical \textbf{V}isual  \textbf{P}refix fusion \textbf{NeT}work (\textbf{HVPNeT})  for visual-enhanced entity and relation extraction.
Specifically, inspired by SimVLM~\cite{SimVLM}, 
we propose visual prefix-guided fusion mechanism
involving \textbf{concatenate} object-level visual representation as the prefix of each self-attention layer in BERT, which is  a more soft and robust attention module for visual enhanced NER and RE. 
We further design a dynamic gate for each layer to generate image-dependent paths, so that a variety of aggregated hierarchical multi-scaled visual features can be considered as visual prefix for enhancing NER and RE.
Overall, we summerize the major contributions of our paper as follows:
\begin{itemize}
    \item We present a hierarchical visual prefix fusion network towards MNER and MRE, incorporating hierarchical multi-scaled visual features through visual prefix-based attention mechanism at each self-attention layer of BERT to generate effective and robust textual representation for reducing error sensitivity.

    \item 
    We utilize the exploitation of dynamic gates to fully leverage the hierarchical visual features. 
    Thus, textual representation of each layer in  Transformer can be aware of  corresponding  hierarchical visual features adaptively. To the best of our knowledge, this paper is the first work to leverage hierarchical pyramidal visual features for multimodal learning.

    \item 
    We evaluate our method on MNER and MRE tasks. Our experimental results on three benchmark datasets validate the effectiveness and superiority of our {\ours}
 
\end{itemize}

\section{Related work}

\textbf{Multimodal Entity and Relation Extraction} 
As the crucial components of information extraction, named entity recognition (NER) and relation extraction (RE) have attracted much attention in the research community~\cite{gcdt,docunet,DBLP:conf/naacl/LiuFTCZHG21,knowprompt,lighter}. 
Previous studies typically focus on textual modality and standard text.
As multimodal data become increasingly popular on social media platforms, early research focusing on textual modality and standard text is limited.
Recently, several studies have focused on the MNER and MRE task, aiming  to utilize the associate images to recognize the named entities and their relation better.

In the early stages, 
\citet{zhang2018adaptive},\citet{lu2018visual}, \cite{moon2018multimodal} and ~\citet{arshad2019aiding} propose to encode the text through RNN and the whole image through CNN, then designing implicit interaction to model information between two modalities to explore multimodal NER tasks.
Recently, \citet{yu-etal-2020-improving,zhang-UMGF} propose to leverage regional image features to represent objects in the image to exploit ﬁne-grained semantic correspondences based on Transformer and visual backbones.

While most of the current methods ignore the error sensitivity, one exception is that \citet{sun2021rpbert}, which proposes to learn a text-image relation classifier to enhance multimodal BERT to reduce the interference from irrelevant images while requiring extensive annotation for the irrelevance of image-text pairs.

\smallskip
\noindent

\textbf{Pre-trained Multimodal Representation} 
The pre-trained multimodal BERT has recently achieved significant improvements in many multimodal tasks (e.g., visual question answering). 
We summarize and compare The existing visual-linguistic BERT models can be divided into two aspects as follows: 
1) \textbf{Architecture}. The single-stream structures consist of Unicoder-VL~\cite{li2020unicoder}, VisualBERT~\cite{li2019visualbert}, VL-BERT~\cite{su2019vl}, and UNITER~\cite{chen2020uniter}, where the text tokens and images are combined into a sequence and fed into BERT to learn contextual embeddings. The two-streams structures, LXMERT~\cite{tan2019lxmert} and ViLBERT~\cite{lu2019vilbert}, separately process the visual and language into two streams with interacting through cross-modality or co-attentional transformer layers.
2) \textbf{Pretraining tasks}. 
The pretraining tasks of multimodal visual-language model mainly consist of masked language modeling (MLM), masked region classification (MRC), and image-text matching (ITM).
However, most of previous models are pre-trained on the datasets of image captioning~\cite{sharma-etal-2018-conceptual,chen2015microsoft} or visual question answering where multimodal interactions are required.  
Applying current visual-language models to the MNER and MRE task may not result in a good performance, since \textbf{MNER and MRE mainly focus on leveraging visual information to enhance the text rather than conducting prediction on the image side}.

\smallskip
 
\section{Methodology}
\begin{figure*}[!htbp] 
\centering 
\includegraphics[width=0.98\textwidth]{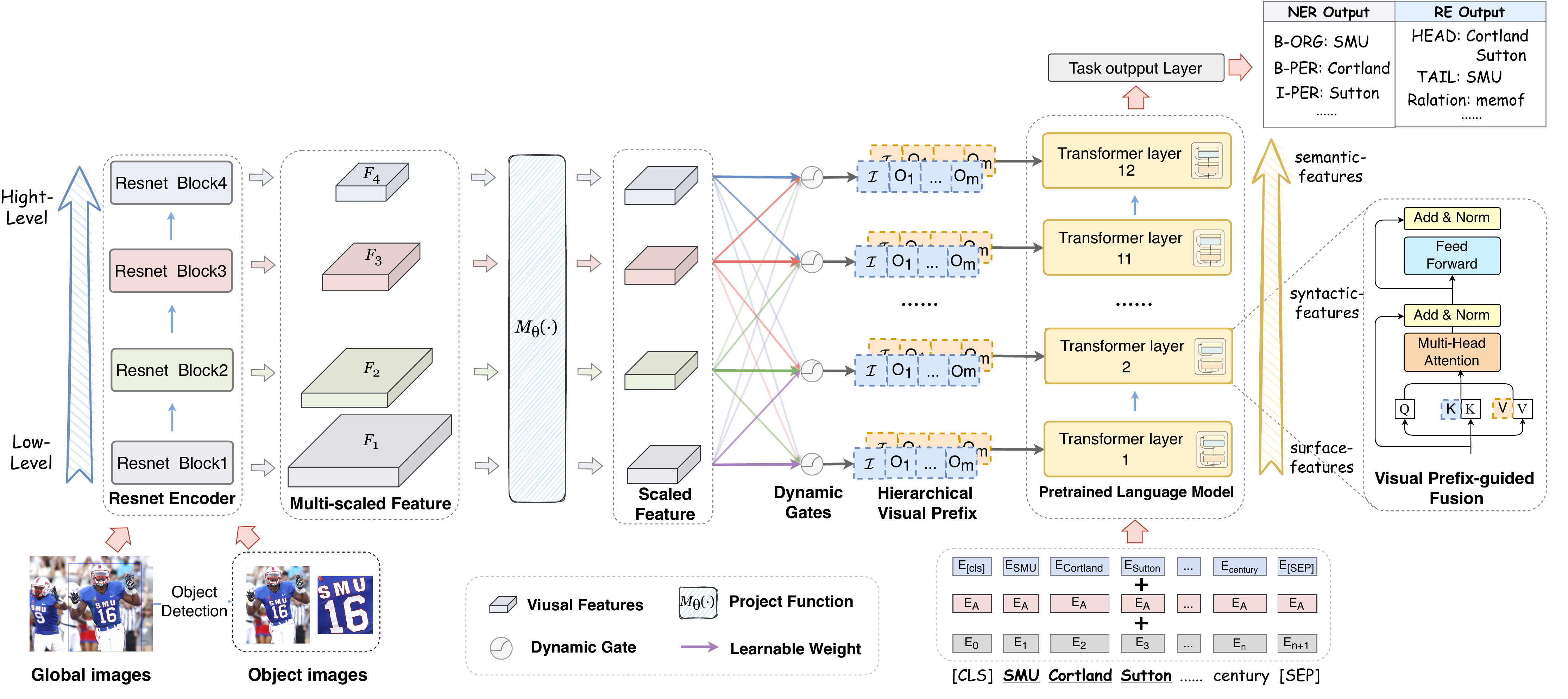} 
\caption{The overall architecture of our hierarchical visual prefix for multimodal entity and relation extraction.} 
\label{fig:model}
\end{figure*}

As illustrated in Figure ~\ref{fig:model}, we present a novel hierarchical prefix fusion  network for multi-modal entity and relation extraction. 
Note that our method can also be applied to other visual-enhanced  tasks towards text.

\subsection{Collection of Pyramidal Visual Feature}
\label{sec:pyramid}
 
On the one hand, the image associated with a sentence maintains several visual objects related to the entities in the sentence, further providing more semantic knowledge to assist information extraction.
On the other hand, the global image features may express abstract concepts, which play the role of a weak learning signal.
Thus, we collect multiple visual clues for multimodal entity and relation extraction, 
which involves taking the regional image as the vital information and the global images as the supplement.
 
Given an image, we follow ~\cite{zhang-UMGF}  to adopt the visual grounding toolkit~\cite{DBLP:conf/iccv/YangGWHYL19} for extracting local visual objects with top $m$ salience.
Then, we  rescale the global image and object image to $224 \times 224$ pixels as the \textbf{global image} $\mathcal{I}$ and \textbf{visual objects} ${\mathcal{O}}=\{o_{1}, o_{2},..., o_{m}, \}$.

In the area of CV, the feature fusion method that leveraging features from different blocks of pre-trained models~\cite{wang2019learning,kim2018parallel,lin2017feature}  is widely  applied for improving model performance. 
Inspired by such practices, we take the first step to focus on the application of pyramid features in the area of multi-modality.
We propose to fuse hierarchical image features into each Transformer layer; thus, leveraging a feature pyramid is essential.
Typically, given an image, we encode it with a backbone model and  generate a list of \textbf{pyramidal feature maps} $\{F_1, F_2, F_3, \dots, F_c\}$ with different scales, then map them with $M_{\theta}(\cdot)$ as follows:
	\begin{align}
	\small
		V_{c} = & \textit{Conv}_{1\times1}(F_c),\\
		V_i = & \textit{Conv}_{1\times1}(\textit{Pool}(F_i)), \ i=1,2,\dot\, c-1,
	\end{align}
where $i$ denotes the $i$-th block of the backbone model, $c$ denotes the number of blocks in the visual backbone model (here is 4 for ResNet), $\textit{Pool}$ represents the pooling operation, where the features are aggregated to the same spatial sizes.
The 1×1 convolutional layer is leveraged to
map the pyramidal visual features to match the embedding size of the Transformer.

\subsection{ Dynamic Gated  Aggregation}
\label{sec:gate}
 
Although objects of different sizes can have appropriate feature representations at the corresponding scales,
it is not trivial to decide which block in the visual backbone is assigned visual prefix for each layer in Transformer.
To address this challenge, we propose constructing the densely connected routing space, where hierarchical multi-scaled visual features are connected with each transformer layer.

\subsubsection{Dynamic Gate Module}

We conduct routine processes through a dynamic gate module, which can be viewed as a procedure of path decision. 
The motivation of the dynamic gate is to predict a normalized vector, which represents how much to execute the visual feature of each block. In the dynamic gate, $g_i^{(l)} \in [0,1]$ denotes the path probability from the 
$i$-th block of visual backbone to the $l$-th layer of Transformer.
It is calculated as $g^{(l)}=\mathbb{G}^{(l)}(V) \in \mathbb{R}^c$, where $\mathbb{G}^{(l)}(\cdot)$ denotes the gating function according to the $l$-th layer in Transformer, $c$ represents the numbers of the block in backbone.
We first produces the logits $\alpha_i^{(l)} $ of the gate signals:
	\begin{equation}
	\small
		\label{FCGate}
		\alpha^{(l)} = f(W_{l}(\frac{1}{c} \sum_{i=1}^c P(V_i))),
	\end{equation}
where $f(\cdot)$ denotes the activate function {Leaky\_ReLU}, $P$ represents the global average pooling layer. 
We first squeeze the input features $V_i$ with a shape of $(d_i, h_i, w)$ from the $i$-th bloc by an average pooling operation. Then we add the features from multiple blocks to generate the average vectors.
We further reduce the feature dimension by $c$ with the  MLP layer $W_{l}$ and consider a soft gate via generating continuous values as path probabilities.
Afterward, we generate the probability vector $g^{(l)}$  for the $l$-th layer  of Transformer as follows:
	\begin{align}
			g^{(l)} = \textit{Softmax}(\alpha^{(l)})
		\label{eq:gumbelsoftmax}
	\end{align}

\subsubsection{Aggregated Hierarchical Feature}
Based on the above dynamic gate $g^{(l)}$, we can derive the final aggregated hierarchical visual feature 
$V_{gated}$ to match the  $l$-th layer in Transformer, as:
\begin{equation}
V_{gated}^{(l)} = g^{(l)} V^{(l)}.\\
\end{equation}
Formally, the final visual features $\tilde{V}_{gated}^{(l)}$ corresponding to  the $l$-th layer of Transformer is obtained by the following concatnation operation,
\begin{equation}
\small
\tilde{V}_{gated}^{(l)}=  [V_{gated}^{(l,I)};V_{gated}^{(l,o_1)};\dots; V_{gated}^{(l,o_m)}],
\end{equation}
which will be adopted  to enhance layer-level representations of textual modality through visual prefix-based attention.

\subsection{Visual Prefix-guided Fusion }
\label{sec:alignment}

We regard hierarchical multi-scaled image feature as visual prefix, and prepend the sequence of visual prefix to the text sequence at each self-attention layer of BERT\cite{devlin2018bert}
In particular, given an input sequence  $X=\{x_1, x_2, ..., x_n\}$, 
the  contextual representations $\boldsymbol{H}^{l-1}\in\mathbb{R}^{n \times d}$ is first projected into the query/key/value vector:
\begin{equation}
\small
\boldsymbol{Q}^l=\boldsymbol{H}^{l-1}\boldsymbol{W}_l^Q, \boldsymbol{K}^l=\boldsymbol{H}^{l-1}\boldsymbol{W}_l^K, \boldsymbol{V}^l=\boldsymbol{H}^{l-1}\boldsymbol{W}_l^V.
\end{equation}
As for aggregated hierarchical visual features $\tilde{V}_{gated}^{(l)}$, we use a set of linear transformations $W_l^{\phi}  \in \mathbb{R}^{d \times 2 \times d}$ for $l$-th layer to project them into the same embedding space\footnote{
 Remarkably, the key and value in the self-attention module contain the different information in two types of semantic space, here $2$ means that we apply two sets of transformation parameters to project aggregated visual features to match the state update process, respectively.} of textual representation in self-attention module.  
Besides, we define the operation of visual prompt $\phi_k^l,\phi_v^l \in \mathbb{R}^{ hw(m+1) \times d}$ as:
\begin{equation}
\small
\{ \phi_k^l, \phi_v^l \}=\tilde{V}_{gated}^{(l)} W_l^{\phi},
\end{equation}
where  $hw(m+1)$  represents the length of the visual sequences, $m$ denotes the number of visual objects detected by the object detection algorithm.
Formally, the visual prefix-based attention are calculated as follows:
\begin{equation}
\small
Prefix\_Attention^{l}={softmax}(\frac{\boldsymbol{Q}^l[{\boldsymbol{\phi}_k^l;\boldsymbol{K}^l}]^T}{\sqrt{d}})
[\boldsymbol{\phi}_v^l;\boldsymbol{V}^l].
\end{equation}

\begin{remark}
\label{rmk:late_interaction_loss}
We regard hierarchical multi-scaled visual features as visual prefix at each fusion layer and sequentially conduct multi-modal attention to update all textual states. 
In this way, the final textual states encode both the context and the cross-modal semantic information simultaneously. which is beneficial to reduce error sensitivity for irrelevant object elements.

\end{remark}

\subsection{Classifier}
Based on above description, we get the final representation of BERT, ${H}^L=U(X,\tilde{V}_{gated}^{(l)})$, where $U(\cdot)$ denotes the operation of visual prefix-based attention.
Finally, we conduct different classifier layers for NER and RE, respectively.
\paragraph{Named Entity Recognition.}
Following ~\cite{moon2018multimodal,yu-etal-2020-improving},
we also adopt the CRF decoder to perform the NER task. 
Formally, we feed the final hidden vectors $\boldsymbol{H}^L=$ of BERT to the CRF model.
For a sequence of tags $y= \{y_1,\ldots,y_n\}$, the probability of the label sequence $y$ and the objective of NER are defined as follows~\cite{lample2016neural}:
\begin{equation}
\label{eq:CRFprob}
\small
\begin{aligned}
\small
p(y|{H}^L) &= \frac{\prod_{i=1}^n{S_i(y_{i-1},y_i,{H}^L)}}{\sum_{y'\in Y} \prod_{i=1}^n{S_i(y'_{i-1},y'_i,{H}^L)}},\\
\mathcal L_{ner} &=-\sum^{M}_{i=1} log(p(y^{(i)}|U(X^{(i)},\tilde{V}_{gated}))).
\end{aligned}
\end{equation}

where $Y$ represents the pre-deﬁned label set with the BIO tagging schema,
and $S(\cdot)$ represents potential functions.
Details can be referred in~\cite{lample2016neural}.

\paragraph{Relation Extraction.}
An RE dataset can be denoted as 
$\mathcal D_{re}=\{(X^{(i)},r^{(i)})\}^M_{i=1}$, 
the goal of RE is to predict the relation $r \in \mathcal{Y}$ between subject entity and object entity.
Specifically, a $\texttt{[CLS]}$ head is utilized to compute the probability distribution over the class set $\mathcal{Y}$ with the softmax function $p(r|X) = \texttt{Softmax}(\mathbf{W}\mathbf{{H}^L}_{\texttt{[CLS]}})$, 
and the parameters of $\mathcal{L}$ and $\mathbf{W}$ are fine-tuned by minimizing the cross-entropy loss over $p(r|X)$ on the entire $\mathcal{X}$ as follows:
\begin{equation}
\label{eq:CRFloss}
\small
\mathcal L_{re} =-\sum^{M}_{i=1} log(p(r^{(i)}|U(X^{(i)},\tilde{V}_{gated}))).
\end{equation}

\section{Experiments}
In the following section, we conduct experiments to evaluate our method on two multimodal information extraction tasks, MNER and MRE. Specifically, we adopt ResNet50~\cite{resnet} as visual backbone and  BERT-base~\cite{devlin2018bert} as textual encoder.
Results on three datasets demonstrate that our {\ours} outperforms a number of unimodal and multimodal approaches.

\subsection{Datasets}
We select three datasets for our experiments: Twitter-2015~\cite{zhang2018adaptive} and Twitter-2017~\cite{lu2018visual}  for MNER, MNRE~\cite{multimodal-re}
for MRE.
Statistical details of datasets  and 
experimental details are provided in Appendix~\ref{sup:data_analysis}, ~\ref{sup:details}.

\begin{table*}[t!]
\centering
\small
\scalebox{0.86}{
\begin{tabular}{c|l|ccc|ccc|ccc}
\hline
\toprule

{\multirow{2}{*}{Modality}} 
& {\multirow{2}{*}{Methods}} 
& \multicolumn{3}{c|}{\textit{Twitter-2015}}
& \multicolumn{3}{c|}{\textit{Twitter-2017}}
& \multicolumn{3}{c}{\textit{MNRE}}
\\
\cmidrule{3-11}
 &  & Precision  & Recall  & F1 & Precision  & Recall  & F1  & Precision  & Recall  & F1 \\

\midrule

    \multirow{6}{*}{Text} 
    & CNN-BiLSTM-CRF~\  & 66.24 & 68.09 & 67.15 & 80.00 & 78.76 & 79.37  & - & - & -
    \\
    & HBiLSTM-CRF~\  & 70.32 & 68.05 & 69.17 & 82.69 & 78.16 & 80.37  & - & - & -
    \\
    & BERT-CRF & 69.22 & 74.59 & 71.81 & 83.32 & 83.57 & 83.44  & - & - & -
    \\
    
    & PCNN~\ & - & - & - & - & - & - & 62.85 & 49.69 & 55.49 
    \\
    & MTB  & - & - & - & - & - & - & 64.46 & 57.81 & 60.86  
    \\

\midrule
    \multirow{12}{*}{Text+Image} 
    & AdapCoAtt-BERT-CRF & 69.87 & 74.59 & 72.15 & 85.13 & 83.20 & 84.10 & - & - & -
    \\
    & OCSGA & \textbf{74.71} & 71.21 & 72.92 & - & - & - & - & - & -
    \\
    & UMT & 71.67 & 75.23 & 73.41 & 85.28 & 85.34 & 85.31  & 62.93 & 63.88 & 63.46
    \\
    & UMGF  & 74.49 & 75.21 & 74.85 & 86.54 & 84.50 & 85.51 & 64.38 & 66.23 & 65.29\\
    
    & BERT+SG & - & - & - & - & - & - & 62.95 & 62.65 & 62.80 
    \\
    & MEGA & 70.35 & 74.58 & 72.35 & 84.03 & 84.75 & 84.39 & 64.51 & 68.44 & 66.41  
    \\
     & VisualBERT  & 68.84 & 71.39 & 70.09 & 84.06 & 85.39 & 84.72  
     & 57.15 & 59.48  & 58.30 
     \\
\cmidrule{2-11}

    
    & \ours-Flat  & 73.76 & 75.32 & 74.54 & 84.43 & 86.42 & 85.41  & 79.32 & 78.20 & 78.75
    \\
    
    & {\ours}-1T3   & 74.25 & 75.45 & 74.85 & 85.43 & 85.85 & 85.75  & 81.18 & 78.46 & 79.25
    \\
    & {\ours}-OnlyObj   & 74.07  & 76.23 & 75.13  & 85.58 & 87.52 & 86.55 & 81.57 & 80.94 & 81.25
    \\
    & \textbf{\ours} 
    & 73.87  & \textbf{76.82}  & \textbf{75.32} & \textbf{85.84} & \textbf{87.93} & \textbf{86.87}
     & \textbf{83.64}  & \textbf{80.78} & \textbf{81.85}
    \\
\bottomrule
\hline
\end{tabular}
}
\caption{\label{tab:result1}
Performance comparison of different competitive baseline approaches for NER and RE. Since the original results of UMT, UMGF and MEGA only involve single extraction task, we reproduce their public code for more comprehensive comparision.
}
\end{table*}

\smallskip
\noindent

\subsection{Compared Baselines}
We compare our {\ours} with several baseline models for a comprehensive comparison to demonstrate the superiority
of our {\ours}. Our comparison mainly focuses on three groups of models: the text-based models, previous SOTA MNER and MRE models, and the variants of our models.
\paragraph{Text-based models:}
we first consider a group of representative text-based  models: 1) {\it CNN-BiLSTM-CRF} ~\cite{acl-MaH16},
2) {\it HBiLSTM-CRF}~\cite{naacl-LampleBSKD16} and
3) \textit{BERT-CRF} for NER.
The following models are specific for RE:
4) \textit{PCNN}~\cite{PCNN};
5) \textit{MTB}~\cite{MTB} is an RE-oriented pretraining model based on BERT.

\paragraph{Previous SOTA models:} besides, we further consider another group of previous SOTA multi-modal approaches for MNER and MRE: 
1) {\it AdapCoAtt-BERT-CRF}~\cite{zhang2018adaptive};
2) {\it OCSGA}~\cite{WuZCCL020};
3) {\it UMT}~\cite{yu-etal-2020-improving};
4) {\it UMGF}~\cite{zhang-UMGF}, the newest SOTA for MNER, which proposes a unified multi-modal graph fusion approach for MNER.
5) \textit{BERT+SG} is proposed in \citet{multimodal-re} for MRE,  which concatenate the textual representation from BERT with visual features generated with scene graph (SG) tool~\cite{sg}.
6) {\it MEGA}~\cite{multimodal-re}, the newest SOTA for MRE, which develops a dual graph for multi-modal alignment to capture this correlation between entities and objects for better performance.
7) \textit{VisualBERT}\citep{li2019visualbert}, different from the above SOTA methods mainly based on co-attention, VisualBERT is a single-stream structure, which is a strong baseline for comparison. 
And the results of VisualBERT listed in our paper are referred from \citet{captionner}

\paragraph{Variants of Our Model:} we set the ablation experiments to explore the effectiveness of our design.
We conduct on the same parameter settings of {\ours} for each variant model for a fair comparison.

\textbf{{\ours}-Flat:} This is another variant of our model without the pyramid structure. Here we assign the visual features with the output of the 4-th block of ResNet and then map the visual features to each layer corresponding to BERT to conduct image-text fusion.

\textbf{{\ours}-1T3:} As ResNet and BERT have four blocks and 12 layers, respectively thus, it is intuitive to directly map visual features in one block to the three layers in BERT. We denote this variant as {\it {\ours}-1T3} to compare with our final version with hierarchical visual features.

\textbf{{\ours}-OnlyObj:} Visual objects are considered as fine-grained image representations. We conduct ablation by only adopting the object-level features in this model to validate the effect of the object features.

\subsection{Overall Performance Comparison}

\subsubsection{Main Results}
The experimental results of {\ours} and all baselines on three testing sets are presented in Table~\ref{tab:result1}.
From the experimental results, we can observe that:

Firstly, we can find that incorporating the visual features is generally helpful for NER and RE tasks by comparing the SOTA multimodal approaches with their corresponding text-based baselines.
Despite previous multimodal approaches can generally achieve better performance, the enormous improvement of F1 score for NER is only about 2.0\% 
(compare UMGF with BERT-CRF), which for RE is about 5.55\% (compare MEGA with MTB). This observation reveals that the performance improvement of images on text-based NER tasks is relatively limited compared with RE tasks.

Secondly, our method is superior to the newest SOTA models UMGF and MEGA, which improves  1.36\%, and 15.44\% F1 scores for Twitter-2017, and MNRE datasets, respectively.  It is worth noting that most of previous multimodal methods ignore the error sensitivity of irrelevant object-level images, while our method regard hierarchical visual prefix  as a prompt for text.
This results indicate that our method can effectively alleviate the error sensitivity irrelevant object images, which is a more robust method for visual enhanced NER and RE.

Finally, we also compare with VisualBERT, which is a pre-trained multimodal BERT with a single-stream structure.
We notice that even as the pre-trained multimodal model, VisualBERT leaves much to be desired in MNER and MRE tasks, which performs worse than UMGF and MEGA, let alone our methods.
We hold that VisualBERT is truly dissatisfactory since the datasets and pre-training process are less relevant to information extraction tasks.

\begin{table*}[t!]
\Large
\resizebox{\linewidth}{!}{
\begin{tabular}[t]{p{8.5cm}p{8.2cm}p{8.0cm}}
\toprule[2pt]
    Relevant Image-text Pair
    & Weak Relevant Image-text Pair
    & Irrelevant Image-text Pair
\\
\midrule[1pt]

\colorbox{red!30}{Taylor Hill} holding \colorbox{yellow!30}{Jun}'s GQ japan lol. 
&
Cold front over \colorbox{yellow!30}{Blyde River Canyon} in \colorbox{red!30}{Limpopo Province}, South Africa.
&
President \colorbox{red!30}{Bush} when he sees the
lights of \colorbox{yellow!30}{America}.
\\
\textbf{Text-Images Attention of {\ours}} \\
\raisebox{-0.94\totalheight}{
    \includegraphics[width=0.96\linewidth]{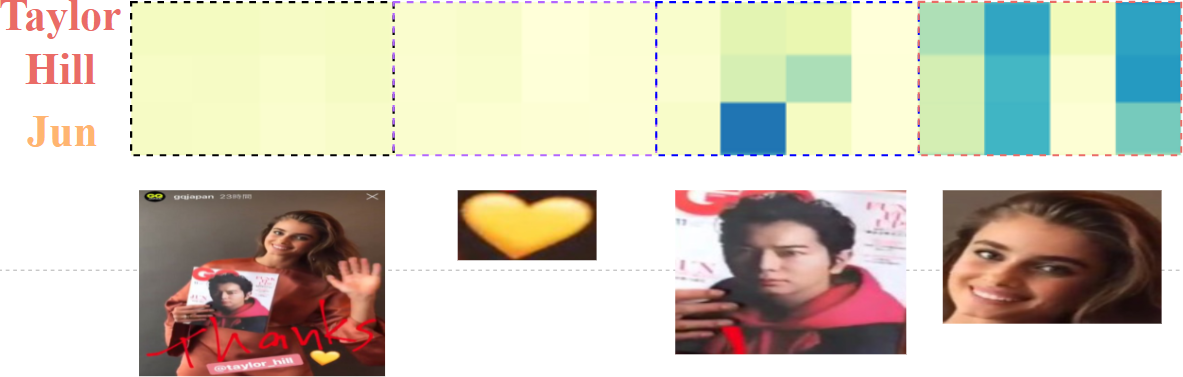} 
} & 
\raisebox{-0.94\totalheight}{
    \includegraphics[width=0.96\linewidth]{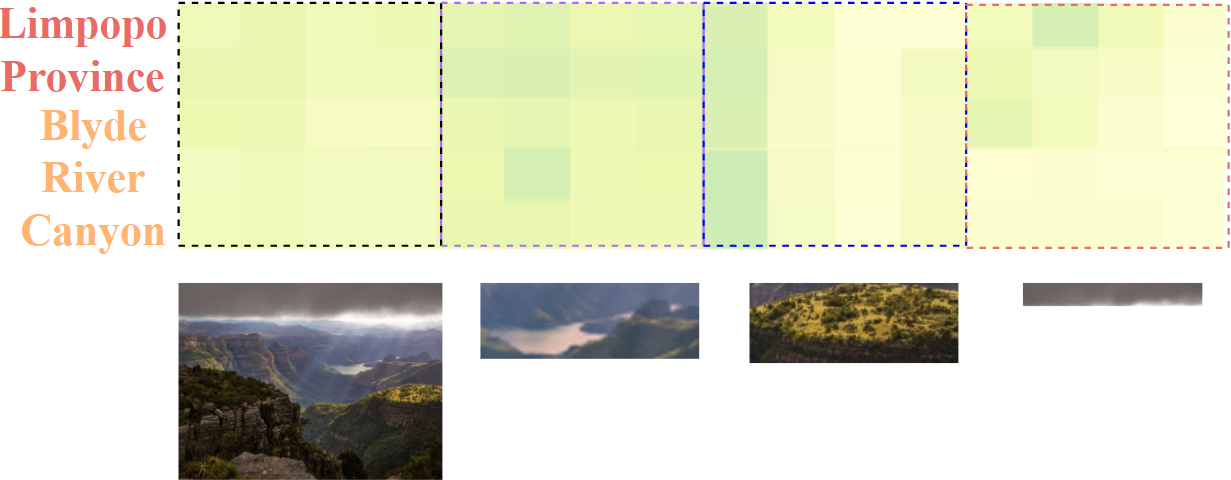} 
}
 & 
\raisebox{-0.94\totalheight}{
    \includegraphics[width=0.97\linewidth]{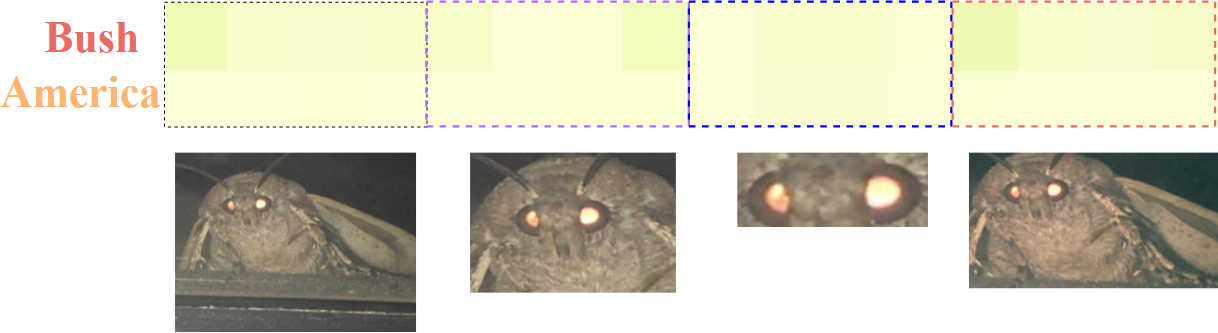} 
}
\\
\textbf{Gold Relations:} 
 per/per/couple & loc/loc/contain & per/loc/place\_of\_residence
\\
\midrule[1pt]

\renewcommand\arraystretch{0.7}
\begin{tabular}[t]{lll}
    BERT:  & \colorbox{red!30}{per}/\colorbox{yellow!30}{per}/couple & \xmark\\
    VisualBERT: & \colorbox{red!30}{per}/\colorbox{yellow!30}{per}/peer & \ding{52}\\
    MEGA: & \colorbox{red!30}{per}/\colorbox{yellow!30}{per}/peer & \ding{52}\\
    {\ours}(Ours): & \colorbox{red!30}{per}/\colorbox{yellow!30}{per}/peer & \ding{52}
\end{tabular} &

\renewcommand\arraystretch{0.7}
\begin{tabular}[t]{lll}
     \colorbox{red!30}{misc}/\colorbox{yellow!30}{misc}/part\_of & \xmark &\\
    \colorbox{red!30}{misc}/\colorbox{yellow!30}{misc}/part\_of & \xmark &\\
    \colorbox{red!30}{per}/\colorbox{yellow!30}{per}/peer & \xmark &\\
     \colorbox{red!30}{loc}/\colorbox{yellow!30}{loc}/contain & \ding{52} &
\end{tabular} &

\renewcommand\arraystretch{0.7}
\begin{tabular}[t]{lll}
     \colorbox{red!30}{per}/\colorbox{yellow!30}{loc}/place\_of\_residence & \ding{52} &\\
    \colorbox{red!30}{misc}/\colorbox{yellow!30}{loc}/held\_on & \xmark &\\
    \colorbox{red!30}{misc}/\colorbox{yellow!30}{loc}/held\_on & \xmark &\\
     \colorbox{red!30}{per}/\colorbox{yellow!30}{loc}/place\_of\_residence & \ding{52} &
\end{tabular} \\

\bottomrule[2pt]
\end{tabular}
}
\caption{The first row shows the split of the relevance of image-text pairs, and the several middle rows indicate representative samples together with their entity-object attention in the test set of MNRE datasets (The y-axis represents the textual entites, and the x-axis denotes the visual objects with length of flattened 4 patches), and the bottom four rows show predicted relation of different approaches on these test samples.
}
\label{tab:re_case}
\end{table*}

\begin{figure}
\hspace{-10pt}
    \centering
    \includegraphics[width=0.23\textwidth]{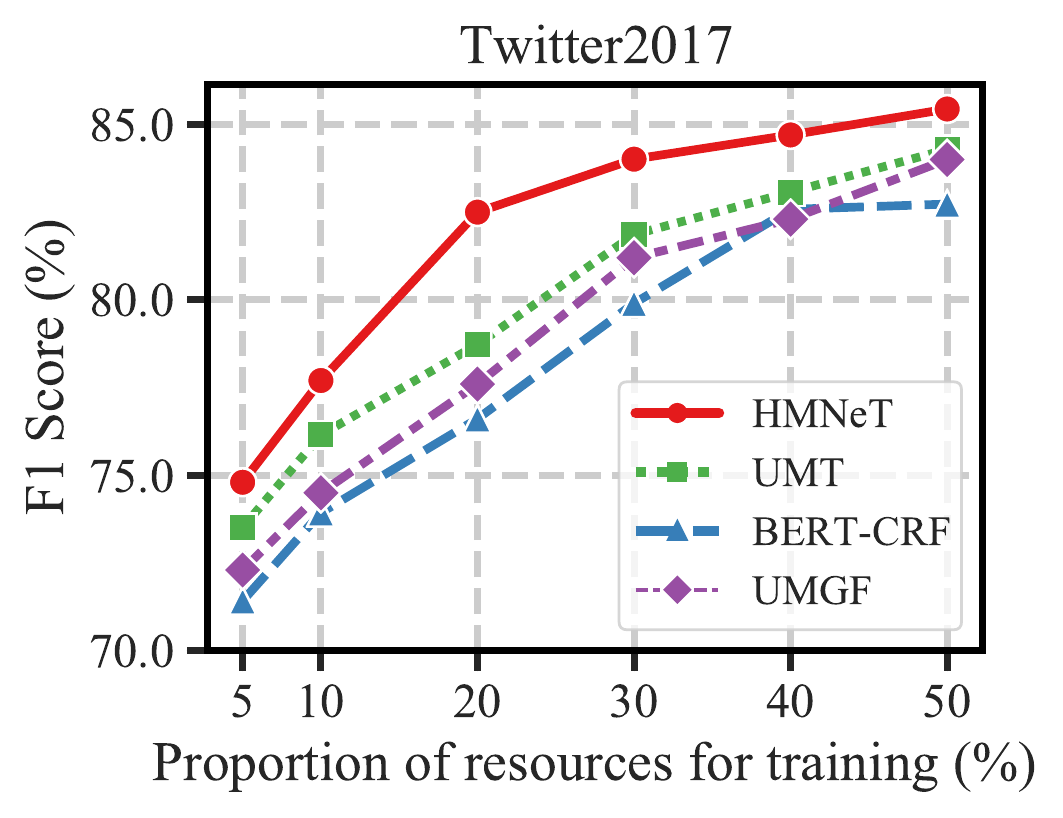}
    \hspace{-20pt}
    \quad
     \includegraphics[width=0.23\textwidth]{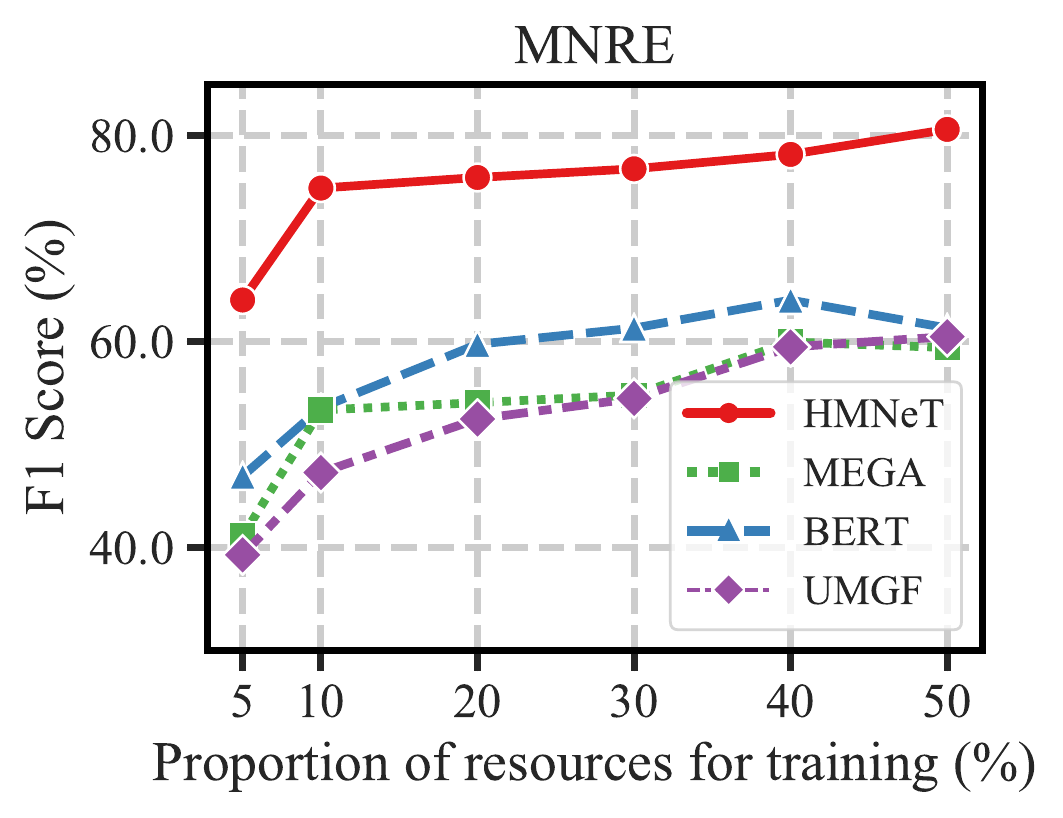}
    \hspace{-20pt}
    \caption{\label{fig:low_resource} Performances on low-resource setting on MNER and MRE task.   
    }
\end{figure}

\subsubsection{Low-resource Scenario}
We further conduct experiments in low-resource settings by randomly sampling 5\% to 50\% from the original training set to form a low-resource training set.
Figure~\ref{fig:low_resource} shows the performance of our method in a low-resource scenario compared with several baselines. 
By analyzing this results, we can observe:
1) UMT and MEGA consistently outperform the compared baselines in the low-resource scenario; the improvement indicates that incorporating the visual features is still helpful for  NER  and  RE tasks in low-resource scenarios.
2) Moreover, it can be observed that the performance of {\ours} still outperforms the other baselines. 
It further proves the effectiveness and data-efficiency of our proposed method.

\subsubsection{Cross-task Scenario}
Table~\ref{tab:cross-task} shows performance comparison of {\ours} and UMGF in a cross-task scenario for versatility analysis. 
For the ﬁrst part, Twitter2017 → MNRE denotes that the trained model on Twitter-2017 is further used to train and test on MNRE. For the second part, MNRE → Twitter-2017 represents that the trained model on Twitter-2017 is used to further train and test on Twitter-2017.
From this Table, we can observe that our {\ours} signiﬁcantly outperforms UMGF by a more considerable margin. 
Note that our method can achieve further improvement in a cross-task scenario, while UMGF performs worse than previous results on the corresponding dataset.
This justifies that our {\ours} is robust to automatically reduce the interference of visual information of irrelevant pictures; thus, more image-text data may facilitate learning better parameters for modality fusion.
Besides, it is also interesting to extend our work to multi-task learning or multi-modal pre-training and we leave these for future works.

\begin{table}[t!]
\centering
\small
\scalebox{0.9}{
\begin{tabular}{l|cc}
\hline
\toprule
{\multirow{1}{*}{Methods}} 

& \multicolumn{1}{c}{\textit{Twitter-2017} $\rightarrow$ \textit{MNRE} }

& \multicolumn{1}{c}{\textit{MNRE} $\rightarrow$ \textit{Twitter-2017}}

\\

\midrule

     UMGF~\  
     & 63.85 $\rightarrow$  62.90 
     \color{blue}{$\downarrow$ (0.95)}
     & 85.51 $\rightarrow$  84.35
     \color{blue}{$\downarrow$ (1.16)}
    \\
    \textbf{\ours}~\  
    & 81.85 $\rightarrow$ 82.50  
    \color{red}{$\uparrow$ (0.75)}
    & 86.87 $\rightarrow$ 87.13
    \color{red}{$\uparrow$ (0.26)}
    \\
\bottomrule
\hline
\end{tabular}
}

\caption{\label{tab:cross-task}
Performance comparison of {\ours} and UMGF in cross-task scenario.}
\end{table}

\subsection{Detailed Model Analysis}

\paragraph{Ablation Study.}
In this part, we conduct extensive experiments with the variants of our model to further analyze the effectiveness of our model.
We illustrate the results of the variant set in Table~\ref{tab:result1} . We can observe that:

\begin{figure*}[!htbp] 
\centering 
\includegraphics[width=1.0\textwidth]{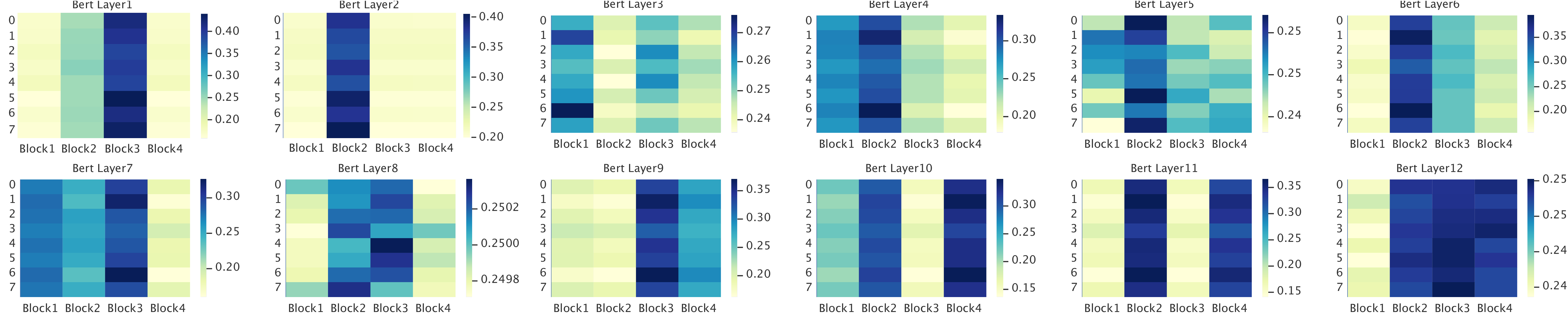} 
\caption{Visualization of dynamic gate learned on MNER task. Each subgraph denotes one layer in BERT, and the ordinate and abscissa respectively represent the instance id in a batch and the block id of ResNet.} 
\label{fig:gate}
\end{figure*}
(1) \textbf{\textit{Visual Prefix-guided Fusion.}}
The core module of our {\ours} is visual prefix-guided fusion, which is a  pluggable operation. Therefore, ablating  visual prefix-guided fusion is equivalent to a purely bert-based baseline model. As shown in Table~\ref{tab:result1}, {\ours} achieve significant improvements over purely bert-based baseline model, revealing the effectiveness of pluggable visual prefix-guided fusion.

(2) \textbf{ \textit{Hierarchical  Visual Features.}} 
To validate the impact of our proposed hierarchical visual features, we carry out experiments by introducing two variants: 1) {\ours}-Flat, crudely assign single visual feature for each layer of BERT; and 2) {\ours}-1T3, intuitively leveraging  visual features from low-level to high-level blocks.
We observe that {\ours} with hierarchical  visual features achieves the best performance consistently compared with the other variants. 
Although the  {\ours}-1T3 performs slightly lower than the version of dynamic gate, it still outperforms the crude variant {\ours}-Flat. 
It reveals that the dynamic gate can automatically learn appropriate weights for multi-scaled visual representations, enabling the model to learn good visual guidance for multimodal entity and relation extraction.

(3) \textbf{ \textit{Visual Clues Term.}} 
As recent SOTA models such as UMT, UMGF, and MEGA all adopt visual objects to enhance textual representation,  we conduct experiments by ablating global images to explore the impact of the visual clues.
As expected, we find that {\ours}-OnlyObj performs slightly worse than {\ours}, which is consistent with the observation of previous works.
This can be attributed to that abstract clues maybe not be associated with the text in information extraction tasks.
In other words, this empirical finding demonstrates the flexibility of our methods to infuse visual clues with different granularity.

\paragraph{Case Analysis for Error Sensitivity}
To validate the effectiveness and robustness of our method, we conduct case analysis for image-text relevance as indicated in Table~\ref{tab:re_case}.
We notice that  VisualBERT, MEGA, and our method can recognize the relation for the relevant image-text pair. We can further find that the attention between relevant entities and objects is significant. While in the situation that image represents the abstract semantic that is weak relevant to the text, only our method success in prediction due to {\ours} captures the more hierarchical features.
It should be noted that another  two multimodal baselines fail in irrelevant image-text pairs while text-based BERT and ours still predict correctly.
These observations reveal that our model regards visual prefix  as a prompt for text may helps learn more robust multimodal representation, which is essential for the noise of uncorrelated object images.

\paragraph{Gate Visualization}
We argue that dynamic gated aggregation for hierarchical visual representation is another 
key component of  {\ours} achieving the superior performance. Specifically, the dynamic gated aggregation can adaptively assign different modality integration paths for different input images, thus, incorporating  visual guidance  with hierarchical multi-scaled information.
To this end, we randomly sample eight images in a batch and visualize their gate vectors learned by {\ours} according to 12 layers of BERT in Figure~\ref{fig:gate}.
Note that optimized gate vectors follow the trend of matching low-level textual semantics with low-level visual semantics and matching high-level textual semantics with high-level visual semantics.
Meanwhile, the modality fusion obtained by dynamic gate learning may provide some valuable insights for efficient visual-language approaches in the future.

\section{Conclusion and Future Work}
In this paper, 
we propose a novel hierarchical visual prefix  fusion neTwork ({\ours}) for visual-enhanced entity and relation extraction. To be specific, we present visual  prefix-guided  fusion by concatenating object-level visual  representation  as  the  prefix  of each self-attention layer in BERT, which is a more soft and robust attention module for visual enhanced NER and RE. We further design leveraging hierarchical multi-scaled visual representation as visual guidance for fusion. 
Intuitively, \textbf{\emph{Good Visual Guidance Makes A Better Extractor}}, and extensive experimental and  results on three benchmarks have demonstrated the effectiveness and robustness of our proposed method.
Meanwhile, our method also face the limitation that it is not suitable for mulimodal tasks in visual side, such as visual grounding.

In the future, we plan to 
1) explore more applications of hierarchical visual prefix in multimodal representation learning, making it more flexible and extensible;
2) try to apply the reverse version of our approach to boost visual representation with text for CV;
3) extend our approach to multitask multimodal pre-training.

\section{Acknowledgments}
We  want to express gratitude to the anonymous reviewers for their hard work and kind comments. 
This work is funded by National Key R\&D Program of China (Funding No.SQ2018YFC000004), NSFC91846204/NSFCU19B2027, Zhejiang Provincial Natural Science Foundation of China (No. LGG22F030011), Ningbo Natural Science Foundation (2021J190), and Yongjiang Talent Introduction Programme (2021A-156-G). 


\bibliography{anthology,custom}
\bibliographystyle{acl_natbib}

\appendix

\section{Detailed Statistics of Dataset}
\label{sup:data_analysis}

\begin{table}[ht]
    \centering
    \small
    \renewcommand{\arraystretch}{1.2}
    \resizebox{0.425\textwidth}{!}{
    \begin{tabular}{l|ccc|c}
        \hline
        \bf Dataset         & \bf Train & \bf Dev   & \bf Test  & \bf \makecell[c]{Avg length \\ (characters)} \\\hline
        Twitter-2015     
        & 4,000 & 1,000   & 3,257 & 95 \\
        Twitter-2017
        & 4,290 & 1,432   & 1,459 & 64 \\
        \hline
    \end{tabular}
    }
    \caption{Size of the datasets in numbers of tweets.}
    \label{tab:datasets1}
\end{table}

\begin{table}[ht]
    \centering
    \small
    \renewcommand{\arraystretch}{1.2}
    \resizebox{0.4\textwidth}{!}{
    \begin{tabular}{l|ccccc}
        \hline
        \bf Dataset         & \bf \# Sent.    & \bf \# Ent.   & \bf \# Rel.   & \bf \# Img.
        
        \\\hline
        TACRED    
        & 53,791 & 152,527   & 41   & - \\
        MNRE
        & 9,201 & 30,970   & 23  & 9,201  \\
        \hline
    \end{tabular}
    }
    \caption{Comparison of MNRE with existing sentence-level Relation Extraction dataset TACRED ( Sent.: sentence, Ent.: entity, Rel.: relation,Img.: image.}
    \label{tab:datasets2}
\end{table}

\section{Experimental Details}
\label{sup:details}
This section details the training procedures and hyperparameters for each of the datasets. We use the BERT-base-uncased model from hugging face library\footnote{https://huggingface.co/}.
We follow UMGF~\cite{zhang-UMGF} to revise some wrong annotations in the Twitter-2015 dataset.
Considering the instability of the few-shot learning, 
we run each experiment 5 times on the random 
seed [1, 49, 1234, 2021, 4321] and report the averaged 
performance. 
We utilize Pytorch to conduct experiments with 1 Nvidia 3090 GPUs. All optimizations are performed with the AdamW optimizer with a linear warmup of learning rate over the first 10\% of gradient updates to a maximum value, then linear decay over the remainder of the training. And weight decay on all non-bias parameters is set to 0.01. 
We set the number of image objects $m$ to 3.
We describe the details of the training hyper-parameters in the following sections.

\subsection{Standard Supervised Setting}
In the MNER task, we fix the batch size as 8 and search for the learning rates in varied intervals [1e-5, 3e-5]. We train the model for 30 epochs and do evaluation after the 16th epoch.  
In the MRE task, we fix the batch size as 32 and learning rates as 1e-5. We train the model for 12 epochs and do evaluation after the 8th epoch.  
In the two tasks, we all choices the model performing the best on the validation set and evaluate it on the test set.

\subsection{Low-Resource Setting}
For different instances per class, we sample five times on the random seed [1, 2, 49, 4321, 1234] and report the averaged performance. For all models, we fix the batch size as 
8 and search for the learning rates in varied intervals [3e-5, 5e-5].  We train the model for 30 epochs and do 
evaluation after the 16th epoch. We choose
the model performing the best on the validation set 
and evaluate it on the test set.

\subsection{Cross-Task Setting}
In the MNER task and RE task, we all use ResNet and BERT-base as the backbone, we transfer the same parameters except the classifier layer and CRF layer when we do cross-task.
In further training, we fix the batch size as  8 and search for the learning rates in varied intervals [1e-5, 3e-5].  We train the model for12 epochs and do evaluation after the 8th epoch. We choose the model performing the best on the validation set and evaluate it on the test set.

\end{document}